\documentclass{article}

\usepackage[nonatbib,final]{neurips_2018}

\usepackage[utf8]{inputenc} 
\usepackage[T1]{fontenc}    
\usepackage[colorlinks=true,linkcolor=blue,citecolor=blue]{hyperref}       
\usepackage{url}            
\usepackage{booktabs}       
\usepackage{amsfonts}       
\usepackage{nicefrac}       
\usepackage{microtype}      
\usepackage{graphicx}
\usepackage{helvet}
\usepackage{courier}
\usepackage{amsmath}
\usepackage{amssymb}
\usepackage{tabularx}
\usepackage{varwidth}
\usepackage{xcolor}
\usepackage{graphicx}
\usepackage{bbm}
\usepackage{bm}
\usepackage{colortbl}
\usepackage[clock]{ifsym}
\usepackage{enumitem}
\usepackage{multicol} 
\usepackage[utf8]{inputenc}
\usepackage{booktabs}  
\usepackage{algorithm}
\usepackage{algorithmic}
\usepackage{wrapfig}
\usepackage{amsthm}
\usepackage{multirow}
\usepackage{subfig}
\usepackage{setspace}
\usepackage{pifont}
\usepackage[export]{adjustbox}

\newcolumntype{C}[1]{>{\centering\arraybackslash}m{#1}}
\newcolumntype{R}[1]{>{\raggedright\arraybackslash}m{#1}}

\usepackage[font=small,skip=10pt]{caption}
\usepackage{setspace}
\DeclareCaptionFormat{myformat}{#1#2#3\hrulefill}

\newcommand{\Cov}{{\textrm{Cov}}}

\title{Deconfounded Score Method: \\ Scoring DAGs with Dense Unobserved Confounding}

\author{
  Alexis Bellot$^{1,2}$\hspace{0.5cm} Mihaela van der Schaar$^{1,2,3}$\\
  $^{1}$University of Cambridge, $^{2}$The Alan Turing Institute, $^{3}$University of California Los Angeles\\
  \texttt{[abellot,mschaar]@turing.ac.uk} \\
}

\begin{document}

\maketitle

\begin{abstract}
Unobserved confounding is one of the greatest challenges for causal discovery. The case in which unobserved variables have a widespread effect on many of the observed ones is particularly difficult because most pairs of variables are conditionally dependent given any other subset, rendering the causal effect unidentifiable. In this paper we show that beyond conditional independencies, under the principle of independent mechanisms, unobserved confounding in this setting leaves a statistical footprint in the observed data distribution that allows for disentangling spurious and causal effects. Using this insight, we demonstrate that a sparse linear Gaussian directed acyclic graph among observed variables may be recovered approximately and propose an adjusted score-based causal discovery algorithm that may be implemented with general purpose solvers and scales to high-dimensional problems. We find, in addition, that despite the conditions we pose to guarantee causal recovery, performance in practice is robust to large deviations in model assumptions.
\end{abstract}

\section{Introduction}
Unmeasured confounding is a long-standing challenge for reliably drawing causal inferences from observational data. This is because, in the presence of unobserved confounding, dependencies observed in data are compatible with many potentially contradictory causal explanations, leaving the scientist unable to distinguish between them \cite{pearl2009causality}.

This paper deals with the discovery of causal relations from a combination of observational data and qualitative assumptions about the nature of causality in the presence of unmeasured confounding. In this scenario, one popular way forward has been to seek an \textit{equivalence class} of mixed graphical models, called maximal ancestral graphs (MAGs) first defined by Richardson et al. \cite{richardson2002ancestral}, including directed, bidirected and undirected edges representing different types of possible causal dependencies compatible with observed conditional independencies. This approach is compelling because it requires no assumptions on the functional relationships between variables or even knowledge on the number or type of unobserved confounders to consistently identify equivalence classes, see e.g. \cite{spirtes2000causation,colombo2012learning,claassen2013learning,triantafillou2016score,tsirlis2018scoring}. 

\begin{figure*}[t]
\centering
\includegraphics[width=0.9\textwidth]{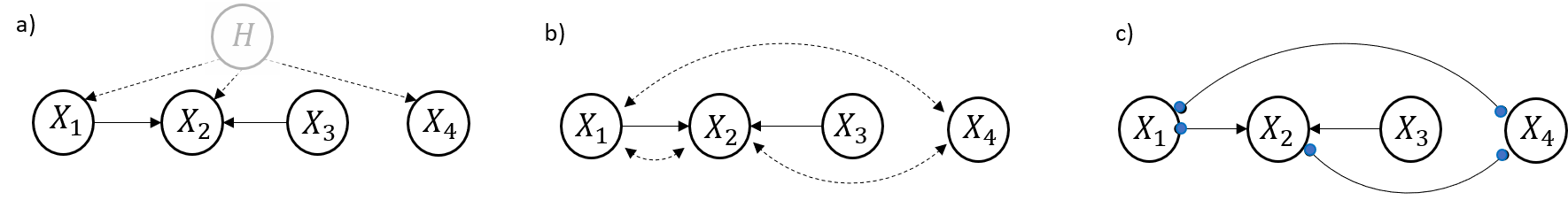}
\caption{MAGs are generally not appropriate to learn causality with dense unobserved confounding. \textbf{a)} A DAG with unobserved confounding $H$ and observed variables $X_1, X_2, X_3$ and $X_4$; \textbf{b)} the corresponding MAG; \textbf{c)} the corresponding equivalence class of MAGs representing the same conditional independences as the DAG (dots \textcolor{blue}{$\bullet$} indicate undetermined causal direction).}
\label{MAG}
\end{figure*}

In some problems however, equivalence classes are largely uninformative as to the underlying causal relationships between observed variables. In genetics for example, as described by \cite{gagnon2013removing,leek2010tackling}, gene expression measurements are often confounded by batch effects, degradation and other specifics of the experiment, leaving most pairs of gene expression measurements conditionally dependent given any subset of other measurements. A similar pattern occurs in finance with asset prices driven by a common political climate or exogenous shocks, even though these events are often not explicitly recorded in data, see e.g. \cite{chamberlain1982arbitrage}. In these examples, graphically, as shown in Figure \ref{MAG}, unobserved confounding when \textit{dense} in its effect on observables (i.e. unobserved variables having an effect on \textit{many} of the observed ones), leaves most edges in the equivalence class of MAGs undetermined.

In this context, we show that we can make progress by restricting ourselves to learning the directed edges among observed variables in a causal MAG (i.e. a directed acyclic graph (DAG)). We study the setting of a high-dimensional system of variables $X\in\mathbb R^p$, in an underlying linear model whose causal interactions are specified by the non-zero entries of a sparse adjacency matrix $W\in\mathbb R^{p\times p}$ encoding the DAG of interest, in the presence of \textit{dense} unobserved confounding $H\in\mathbb R^q$,
\begin{align}
\label{model_intro}
    X = W X + B H + E,
\end{align}
where $E$ is a vector of errors but realizations of $X$ may be confounded by $H$ through $B\in\mathbb R^{p\times q}$.

\subsection{Contributions}
A practical consequence of dense unobserved confounding is that the contributions to the matrix of covariances $\Cov(X)$ of confounding matrix $B$ is \textit{different} (in a characteristic sense) from the contribution due to the matrix of causal contributions $W$. A property that can be used to adjust $X$ for confounded contributions by analogizing DAG learning to a regression problem involving a sparse plus dense or low rank superposition of matrices, studied for example by \cite{candes2011robust,shah2020right, cevid2018spectral}, in this case interpreted as causal and confounded contributions respectively in the context of unobserved confounding.

We show that one can formulate DAG learning among $p$ observed variables in the presence of dense unobserved confounding as the solution of an optimization program:
\begin{align}
\label{score_function}
    \text{minimize}\hspace{0.2cm} \mathcal S(W;\mathbf X) \quad\text{such that}\quad W \in \mathbb D,
\end{align}
where $\mathbb D$ is the set of $p \times p$ matrices representing the weighted adjacency matrix of a DAG and $\mathbf X\in\mathbb R^{n\times p}$ is the data. $\mathcal S$ is known as the score function. Estimators of this form have a long history in causal discovery, see e.g. \cite{aragam2015learning, scutari2018learns, chickering2002learning, zheng2018dags, buhlmann2011statistics,loh2014high}, predominantly in the fully observed setting. Our contributions are three-fold.
\begin{enumerate}[leftmargin=*,itemsep=0pt]
    \item We show that in high-dimensional $(p \gg n)$ linear models (\ref{model_intro}) the spectrum of the confounded data matrix is characteristically different than would be expected without unobserved confounding.
    \item With this insight, we propose a score function $\mathcal S$ and problem (\ref{score_function}) whose solution has explicit finite-sample true positive guarantees.
    \item We develop a practical two-stage algorithm, the Deconfounded Score (DECS) method, leveraging standard gradient-based optimization solvers and algebraic acyclicity formulations of DAGs that has the practical benefits of being much simpler and scaling better to large samples and high-dimensional feature spaces than alternative independence-based approaches.
\end{enumerate}


\subsection{Related work}
\label{sec_work}
This paper primarily engages with the literature on causal discovery in the presence of unobserved confounding but also draws on insights from high-dimensional linear regression and factor models. 

We argue for exploiting properties of the spectrum of data matrix to recover a causal DAG among observed variables in high-dimensional systems. We contrast this approach with work that seeks conditional independencies as a route to causality, first presented in \cite{spirtes2000causation} and subsequently widely extended and applied e.g., \cite{richardson2002ancestral,colombo2012learning,claassen2013learning,colombo2014order}. The authors developed theoretically consistent algorithms for recovering an equivalence class of MAGs which may be linked to the underlying causal structure with an assumption of faithfulness. Examples include the FCI, FCI+, RFCI and other variants that use (a polynomial number of) conditional independence tests to iteratively recover the skeleton and some edge orientations. A second class of algorithms instead propose to search greedily in the space of MAGs optimizing a score function defining goodness of fit on the observed data. For instance, \cite{triantafillou2016score} proposed a greedy search algorithm maximizing a penalized Gaussian likelihood score over the class of MAGs, \cite{bernstein2020ordering} proposed a greedy search over partial orderings of the variables, \cite{frot2017robust} use a decomposition of the covariance matrix into sparse and low rank components before applying the GES algorithm \cite{chickering2002learning}, \cite{tsirlis2018scoring} proposed a hybrid combination of score and independence-based algorithms, among others that consider bow-free acyclic graphs (a special case of MAGs) studied by \cite{nowzohour2015structure,drton2009computing}.

We share the objective of seeking a consistent score function but instead aim to recover a DAG among observed variables only and do so focusing on high-dimensional spaces from a penalized regression perspective, relying instead on the principle of independent mechanisms \cite{janzing2018detecting} to link the spectrum of the data matrix to causality. This challenge is related to the literature on identifiability in high-dimensional regression \cite{chernozhukov2017lava,candes2011robust,shah2020right,cevid2018spectral} and estimation in linear factor models \cite{fan2013large,fan2018large,bai2003inferential,fan2018large,bing2020adaptive}. For instance, in different variations of the underlying factor model it is possible to consistently recover a decomposition of regression parameters or covariance matrices into a sparse component and a dense or low-rank component separately. This paper applies this theory to extend (fully-observed data) score-based DAG learning consistency results (e.g. \cite{aragam2015learning,aragam2019globally}) to a special case of unobserved confounding that could not be consistently analysed before. 

\section{Problem formulation}
\label{sec_formulation}
We use the language of structural causal models as our basic semantical framework, see e.g. \cite{pearl2009causality}. We suppose a structural causal model describes a natural phenomenon of interest, partially observed through a random vector $X = (X_1, \dots , X_p)$ satisfying,
\begin{align}
\label{model}
    X = W X + B H + E,
\end{align}
where $W\in\mathbb D$ is an adjacency matrix that specifies the causal variable relationships among $X$. $H = (H_1, \dots, H_q)$ is a vector of $q$ unobserved Gaussian confounders that influence $X$ through a dense matrix $B\in\mathbb R^{p\times q}$. $E = (E_1,\dots,E_p)$ a vector of independent sources of noise also drawn from a Gaussian distribution. We will assume $p \gg n$ and $p \gg q$.

\textbf{Assumption 1.} (Dense random matrix $B$.) \textit{There exists $K>0$ such that $\sigma_{\min}(B) \geq K\sqrt{p}$ with high probability, i.e. the smallest singular value of $B$ has asymptotically at least the same rate as $\sqrt{p}$, which is satisfied e.g. if each entry in $B$ is Gaussian (see e.g. equation (1.5) in \cite{rudelson2009smallest}) -- i.e. $B$ is dense. }

We will denote $\mathbf X \in\mathbb R^{n \times p}$ the data matrix and $\mathbf H \in\mathbb R^{n \times q}$ the corresponding matrix of realizations of unobserved variables. $W$ defines a DAG over the observed variables: if $[W]_{ij}\neq 0$ we will say that $X_j \in \texttt{Pa}(X_i)$ is a causal parent of $X_i$\footnote{Edges specified by $W$ have the interpretation of a direct causal link between variables in $X$ but the absence of edges does not necessarily reflect a conditional independence, contrary to edges in MAGs.}. Our goal is to define a score function $\mathcal S:\mathbb R^{p\times p} \rightarrow \mathbb R$, involving only the observed data, that provably attains a minimum at the weighted adjacency matrix $W$ of the underlying DAG. 


\subsection{The challenge of high-dimensional data} In high-dimensional systems, defining a function that scores candidate adjacency matrices $W$ is intrinsically ill-posed without further structure. When $\text{Rank}(\mathbf X) < p$, e.g. when $p > n$, there are infinitely many solutions with minimum score. Given one solution $W^{\star}$, the quantity $W^{\star} + \epsilon$ is also a solution for any $\epsilon$ in the null space of $\mathbf X$. Moreover, even if only signs are desired in the underlying DAG (i.e. we seek to know whether each estimated causal effect raises or lowers the probability of outcomes in children nodes), this type of non-uniqueness makes interpretation of solutions cumbersome: for any $i$ and at least one $j \in \{1, \dots, p\}$, we will have $W_{ij}^{\star} > 0$ for one solution, but $W_{ij}^{\star} < 0$ for another solution. Constraining solutions to be sparse (i.e. few edges relative to the number of variables) is one way to overcome this problem \cite{hastie2015statistical}. 

\subsection{The challenge of confounded data} 
An assumption of sparsity on solutions to score-based optimization problems such as (\ref{score_function}) is not appropriate however. The inferred matrix of associations will typically be dense as a result of confounding. We may write for instance,
\begin{align}
\label{perturbed_model}
    X = (W+C) X + (B H - CX) +  E,
\end{align}
where $C\in\mathbb R^{p\times p}$ is chosen such that its $i$-th row, $c_i$ satisfies $\text{Cov}(b_i^T H - c_i^T X,X_i)=0$ for $i=1,\dots,p$. $b_i$ is the $i$-th row of $B$ and $c_{ii}=0$. $C$ is the scaled projection of $H$ on $X$: $c_i = \Cov(X)^{-1}\Cov(X,H)b_i$, and represents the bias introduced in the estimation of $W$ due to the contributions of unobserved confounding variables $H$. If we ignore confounding, we shall have $W+C$ as the target of score-based algorithms instead of $W$. The bias in the estimation of $W$ is potentially large if $||\mathbf Xc_i||_2$ is large for each $i$, where. We rename the error vector of this model $\tilde E := (B H - CX) +  E$, each entry independently distributed and independent of $X$ by construction. 

\section{Adjusted Scoring of DAGs}
In this section, we describe the principle of independent causal mechanisms which motivates an adjusted score function that mitigates the contribution of unobserved confounding while preserving the causality among observed variables.

\subsection{The asymmetry of confounding}
If we were to be given the underlying causal structure and all variables fully observed ($H=0$), in its canonical form $\mathbb P(X_1,\dots, X_p) = \prod_{i=1}^p \mathbb P(X_i |\texttt{Pa}(X_i))$. Under the principle of independent mechanisms, the conditional distributions $\mathbb P(X_i |\texttt{Pa}(X_i))$ have the property of describing an invariant mechanism of nature that should be independent of the distribution of the causes $\mathbb P(\texttt{Pa}(X_i))$, see e.g. \cite{parascandolo2018learning,janzing2018detecting,janzing2018detecting2}. 

Given that the underlying model of variable associations (\ref{model}) is linear we way define this independence criterion by associating each $\mathbb P(X_i |\texttt{Pa}(X_i))$ with the set of parameters $\mathbf w_i$ (i.e. the $i^{th}$ row of $W$) and the distribution of its parents $\mathbb P(\texttt{Pa}(X_i))$ with the matrix of second moments of $X$ that fully specifies the distribution in the Gaussian model (\ref{model}). Following the principle of independent mechanisms, intuitively, each $\mathbf w_i$ should be "independent" from $\Cov(X)$, and specifically, it would be unexpected to find $\mathbf w_i$ aligned in any specific manner to large principal components of $\mathbf X$. In the presence of unobserved confounding this changes since unobserved confounding induces a dependence between $X_i$ and its parents $\texttt{Pa}(X_i)$: the independence of causal mechanisms is not expected to hold and will induce a statistical footprint in the distribution of the observed data that is different than it would be without confounding\footnote{With access to data from different environments, invariances in the presence of unobserved confounders for causal discovery have also been proposed \cite{rothenhausler2019causal,bellot2020accounting}.}. The following Lemma shows that the \textit{direction} of the confounded contribution tends to be concentrated in specific vectors related to the covariance matrix of $X$. 

\textbf{Lemma 1.} \textit{Assume that $H$ is univariate. Then, the principal components of $X$ are approximately aligned with the columns of $B$ and approximately aligned with each row of the confounded contribution $C$.}

\textit{Proof sketch.} $B\in\mathbb R^p$ (i.e. a column vector since $H$ is univariate) in (\ref{model}) tends to be approximately aligned with $\Cov(X) = (I-W)^{-1} (BB^T+\Cov(E)) (I-W)^{-T}$ since $B$  is an eigenvector of $(BB^T+I)$ with large eigenvalue and $W$ is sparse, assuming $\Cov(E)\approx I$. And therefore also the direction of the $i$-th row of the perturbation $C$, $c_i = \Cov(X)^{-1}\Cov(X,H)B$, as a multiple of $B$, must be approximately aligned with large eigenvectors of $\Cov(X)$. 

\subsection{Adjusting for confounding}
The more each row of $B$ is aligned with large singular vectors of $\mathbf X$, the larger $\|\mathbf X b_i\|_2$ will be. Under the principle of independent mechanisms, such alignment between rows of $W$ and large singular values of $\mathbf X$ is unlikely. 

\textbf{Lemma 2.} \textit{Under the principle of independent mechanisms, the rows of $W$ are orthogonal to the principal components of $X$ with high probability.}

\textit{Proof.} In high dimensional systems any two randomly chosen vectors, as would be any pair of rows of $W$ and principal components of $X$ under the principle of independent mechanisms, are  orthogonal with high probability by e.g. Proposition 2.1 in \cite{gorban2018blessing}.

We can expect therefore that shrinking large principal components of $\mathbf X$ shrinks the contribution of $C$ in our estimates but leaves the contribution due to causal coefficients $W$ unchanged as these are largely orthogonal. One practical approach is thus to remove or truncate large singular values of $\mathbf X$ leaving the direction of singular vectors unchanged, as has been proposed in the context of high-dimensional regression with the lava estimator \cite{chernozhukov2017lava}, PCA adjustment techniques, see e.g. \cite{fan2013large} or the trim transform \cite{cevid2018spectral}. Following \cite{cevid2018spectral}, let $\mathbf X = \mathbf U \mathbf D\mathbf V^\intercal$ be the singular value decomposition of $\mathbf X\in\mathbb R^{n\times p}$, where $\mathbf U \in\mathbb R^{n\times r}$, $\mathbf D \in\mathbb R^{r\times r}$, $\mathbf V \in\mathbb R^{p\times r}$, and where $r = \min(n, p)$ is the rank of $\mathbf X$. We write $d_1 \leq d_2 \leq \dots \leq d_r$ for the diagonal elements of $\mathbf D$. We use the truncated form of the singular value decomposition, which uses only non-zero singular values. We define the adjusted matrix $\tilde{\mathbf X} := \mathbf F\mathbf X$ as a transformation of $\mathbf X$ by $\mathbf F \in\mathbb R^{n\times n}$ that upper-bounds each singular value to $\tilde d := \text{median}(d_1,\dots,d_r)$: $\mathbf F:= \mathbf U\tilde{\mathbf D} \mathbf U^\intercal$, where $\tilde{\mathbf D}$ is diagonal with each element on the diagonal equal to $[\tilde{\mathbf D}]_{ii}:= \min(d_i, \tilde d) / d_i$. 

\textbf{Lemma 3.} (Effect of adjustments $\mathbf F$). \textit{Under the model assumptions, $||\mathbf X C||_2 \leq ||\mathbf H B||_2 = \mathcal O(\sqrt{np}\cdot\mathbb E||B_1||_2)$. In contrast, $||\tilde{\mathbf X}C||_2 \leq \mathcal O(\sqrt{p}\cdot\mathbb E||B_1||_2)$.}

\textit{Proof.} $\|\cdot\|_2$ is the operator norm when applied to matrices. This is an application of Lemma 1 in \cite{cevid2018spectral}.


\subsection{An adjusted score function}
A score-based DAG estimator that derives from this approach is immediate, formulated as the solution of a constrained optimization problem,
\begin{align}
\label{optimization}
    \widehat W \in \underset{W \in \mathbb D}{\text{argmin}}\hspace{0.3cm} \mathcal S(W; \mathbf X), \qquad
    \mathcal S(W; \mathbf X):=\frac{1}{2n} || \tilde{\mathbf X} - \tilde{\mathbf X} W ||^2_F + \lambda ||W||_1,
\end{align}
where $\tilde{\mathbf X} = F\mathbf X \in \mathbb R^{n \times p}$ is the linear transformation of the data matrix that truncates large principal components. $\mathcal S$ is the penalized mean squared score function in Frobenius norm, $\lambda ||W||_1$ is the scaled sum of the magnitude of the entries in $W$, and $\lambda >0$.

\subsection{A guarantee on recovery of $W$}
\label{sec_guarantees}
An important question is whether solutions to the adjusted optimization problem in fact converge, and if so, whether they converge to the underlying causal structure $W$. 

The problem in (\ref{optimization}) can be interpreted as optimization over a family of neighbourhood regression problems, each variable regressed on its non-descendants. This decomposition can be used to derive uniform bounds on recovery error. In particular, \cite{aragam2015learning} first showed that imposing sparsity on the true DAG substantially reduces the number of regressions, otherwise equal to $2^{p-1}p$ (since the topological ordering of the DAG, or the set of non-descendants for each variable is unknown a priori) and intractable in general. Penalized score-based learning without unobserved confounding, they showed, efficiently and provably recovers a sparse DAG $W_{\min}$ with minimum conditional variance, also called minimum-trace DAG. If unique $W_{\min}$ equals $W$, otherwise there is technically no truth to approximate from data, though penalized score-based learning does converge to a sparse representative among the class of minimum-trace DAGs. We refer to \cite{aragam2015learning, aragam2019globally} for more details.


In this section, we show that a similar strategy applies in our setting, with the difference however that each neighbourhood regression problem, instead of being a penalized regression problem, is formulated as the following adjusted, penalized regression problem,
\begin{align}
\label{cevic}
    \underset{\mathbf w_i \in\mathbb R^p,\hspace{0.1cm}  \text{supp}(\mathbf w_i)\subset S}{\text{arg min}}\hspace{0.3cm} \frac{1}{2n} || \tilde{\mathbf{X}}_i - \mathbf w_i^T\tilde{\mathbf X}  ||^2_2 + \lambda ||\mathbf w_i||_1.
\end{align}
$S$ is a subset of all variables other that $X_i$ that defines a \textit{neighbourhood} of $X_i$. $\tilde{\mathbf X}_i \in \mathbb R^n$ is the $i^{th}$ column of $\tilde{\mathbf X}$, $\mathbf w_i \in \mathbb R^p$ is the $i^{th}$ column of $W$ (i.e. the regression parameters defining the parents of $X_i$) and supp$(\mathbf w_i)$ denotes the support of $\mathbf w_i$.

To obtain uniform bounds on the error in DAG estimation as in \cite{aragam2015learning,aragam2019globally} it suffices to show that each regression parameter $\mathbf w_i$ can be recovered consistently. Bounds on the estimation of $\mathbf w_i$ (in $l_1$ of $l_2$ norms for example), exist in the high-dimensional regression literature once we recognise $\mathbf w_i$ as the sparse component in a sparse plus dense superposition of regression parameters e.g., Theorem 1 in \cite{cevid2018spectral}. Two conditions are needed for these bounds. First, assumption 1 defined in section \ref{sec_formulation} which formalizes the fact that the effect of $H$ is spread over many observables. Second, we must ensure the transformation $F$ to be well-behaved, i.e. not shrink the causal signal too much (specifically imposing a smallest restricted eigenvalue condition on the covariance matrix of $\tilde X$) but consistently lower large singular vectors of $\tilde X$. We refer to the Appendix for a formal statement of all conditions. 

For any $A \in \mathbb R^{p\times p}$, let $\tau(A) := \min\{|a_{ij}| : a_{ij} \neq 0\}$. The quantity $\tau(W_{\min})$ measures the smallest nonzero weight in $W_{\min}$, which is a measure of the signal strength in the problem. Denote $a \gtrsim b$ to mean that $a \geq C \cdot b$ for some constant $C > 0$, and $\sigma := \max_i(\sigma_i)$ where $\sigma_i$ in the standard deviation of adjusted error terms $\tilde{E_i}F$ ($\tilde{E_i}$ defined at the end of section \ref{sec_formulation}. The following Theorem shows that the support of the minimum-trace DAG, i.e. the true edges in the underlying DAG, is contained in the support of the estimated DAG with high-probability.

\textbf{Theorem 1}. (True positive guarantee) \textit{Under regularity conditions and $W_{\min}$ unique, for $n \gtrsim s \log p$, $\lambda \gtrsim \sigma \sqrt{\log p/n}$, and $\tau(W_{\min}) \gtrsim \lambda$,}
\begin{align}
    \text{supp}(W_{\min}) \subseteq \text{supp}(\widehat W),  
\end{align}
\textit{with probability $1-\mathcal O(e^{k \log p})$, where $k$ is the maximum in-degree of $W_{\min}$, i.e. the maximum number of directed edges that point into any observed node, and $s$ is the size of the support of $W_{\min}$.}

\textit{Proof.} The proof is given in the Appendix.

Despite the presence of unobserved confounding, this results guarantees not to miss any causal edges in the true network but we may (typically) have too many false positive selections in the estimated DAG. This result is equivalent to the property of \textit{variable screening} of the lasso estimator. Results exist also to guarantee \textit{full} support recovery of the lasso estimator \cite{wainwright2009sharp}. In the DAG estimation setting however, this necessitates however much stronger conditions, roughly speaking requiring that no parent of a given variable be highly correlated with "non-parent" variables, known as the incoherence condition discussed by \cite{aragam2019globally}. However, we do demonstrate  empirically at least that our method does have lower false discovery rates than competing approaches in Section \ref{sec_experiments_main}.

\subsection{Practical algorithms}
This section describes a practical algorithm to solve (up to stationarity) the constrained optimization problem (\ref{optimization}). The practical challenge is to enforce efficiently the acyclicity constraint on $W$. One approach is to transform the traditional combinatorial optimization problem into a continuous program, using an equivalent formulation of acyclicity via the trace exponential function, due to \cite{zheng2018dags}. $W$ corresponds to an acyclic graph if and only if the function $h(W)=0$, where $h(W):= Tr(\exp\{W \odot W\}) - p$, $\exp\{M\}$ denotes the matrix exponential of a matrix $M$, $\odot$ denotes the element-wise matrix product, and $Tr$ denotes the matrix trace operator. The optimization problem becomes,
\begin{align}
\label{continuous_optimization}
    \underset{W \in \mathbb R^{d\times d}}{\text{minimize}}\hspace{0.3cm} \frac{1}{2n} || \tilde{\mathbf X} - \tilde{\mathbf X} W ||^2_F + \lambda ||W||_1 \qquad\text{such that} \quad h(W)=0,
\end{align}
which is non-convex but can be solved approximately with second-order methods as done by \cite{zheng2018dags}. We use their augmented Lagrangian method, with resulting solutions shown to be very close to the true global minimum in practice and that scale to modern problem sizes with thousands of variables\footnote{Recently, \cite{ng2020role} found that enforcing $h(W)=0$ may not be necessary to recover a DAG in practice, and argue for a soft constraint leading to faster methods. One may extent the above in the same manner.}.

Choosing the regularization parameter $\lambda$ with cross-validation is different than in the standard setting with no confounding. When using cross-validation, aiming for best prediction, the chosen $\lambda$ would be typically too small since the best prediction would also try to capture the unwanted signal from $\mathbf XC$ in (\ref{perturbed_model}). To partially correct for this issue, cross-validation should be run on the adjusted data $\mathbf{\tilde X}$. 
We call this causal discovery approach the Deconfounded Score method (\textbf{DECS}).

\section{Experiments on synthetic data}
\label{sec_experiments_main}

Our goal in this section is to measure causal discovery performance in extensive experiments, and especially under violations of our assumptions.

\textbf{Comparisons.} We make comparisons with three causal discovery methods: the independence-based Fast Causal Inference (\textbf{FCI}) \cite{spirtes2000causation}, \textbf{LGES} \cite{frot2017robust} that uses a decomposition of the covariance matrix followed by the GES algorithm, and \textbf{Notears} \cite{zheng2018dags}, the continuous optimization approach without adjustments (it is not specifically designed for unobserved confounding but serves to isolate the benefit / harm of adjusting for unobserved confounding with \textbf{DECS}). We note that the performance of non-convex optimization programs in the context of DAG learning, and the benefit of continuous-optimization formulations for DAG learning are well studied \cite{zheng2018dags,ng2020role} -- both noting significant gains over independence-based methods.

\textbf{Metric.} Note however that not all algorithms have the same output, FCI outputs an equivalence class of MAGs, LGES outputs an equivalence class of DAGs, and Notears outputs a weighted adjacency matrix. For consistent performance comparisons, we chose to consider the \textit{skeleton} (i.e. all directionality omitted) of estimated graphs which is a common output across all algorithms. In a sense this treats existing algorithms favourably by regarding undirected or undetermined edges as true positives as long as the true graph has a directed edge in place of the undirected edge. (We give more details on algorithm and metric implementation in the Appendix). We report the AUC and SHD on estimated skeletons and both take into account false positives and false negatives. We do make more detailed evaluations in the Appendix considering the error in weighted adjacency recovery $(W-\hat W)^2$ (although comparisons there are limited to Notears which is the only baseline outputting weighted adjacency matrices).

\subsection{Experimental set-up} 
In each experiment, we generated a $p$-dimensional random graph $G$ from a Erd\"os–R\'enyi random graph model with $p$ edges on average. Given $G$, we assigned uniformly random edge weights to obtain a weighted adjacency matrix $W\in\mathbb R^{p\times p}$. Given $W$, we sampled $X = W X + B H +  E$ repeatedly from different noise models for $H\in \mathbb R^{q}$ and $E\in\mathbb R^p$, including Gaussian, Exponential and Gumbel distributions, and $B \in \mathbb R^{p\times q}$ with each entry independently sampled from $\mathcal N(0, 1)$. We fix the number of observations $n=100$ in all experiments.

\textbf{Task.} The task is to recover the skeleton defined by $W$ (i.e. the matrix $\bar W$ such that $[\bar W]_{ij} = \mathbf 1\{[W]_{ij}\neq 0\}$) given $n$ independent samples from $X$. We consider performance comparisons along the spectrum of five parameters: the data distribution family, the dimensionality $p$ of $X$, the dimensionality $q$ of $H$, the noise scale $\sigma$ which when small implies a more pronounced perturbation of unobserved confounding, the denseness of $B$.

\begin{figure*}%
    \centering
    \subfloat{{\includegraphics[width=7.4cm]{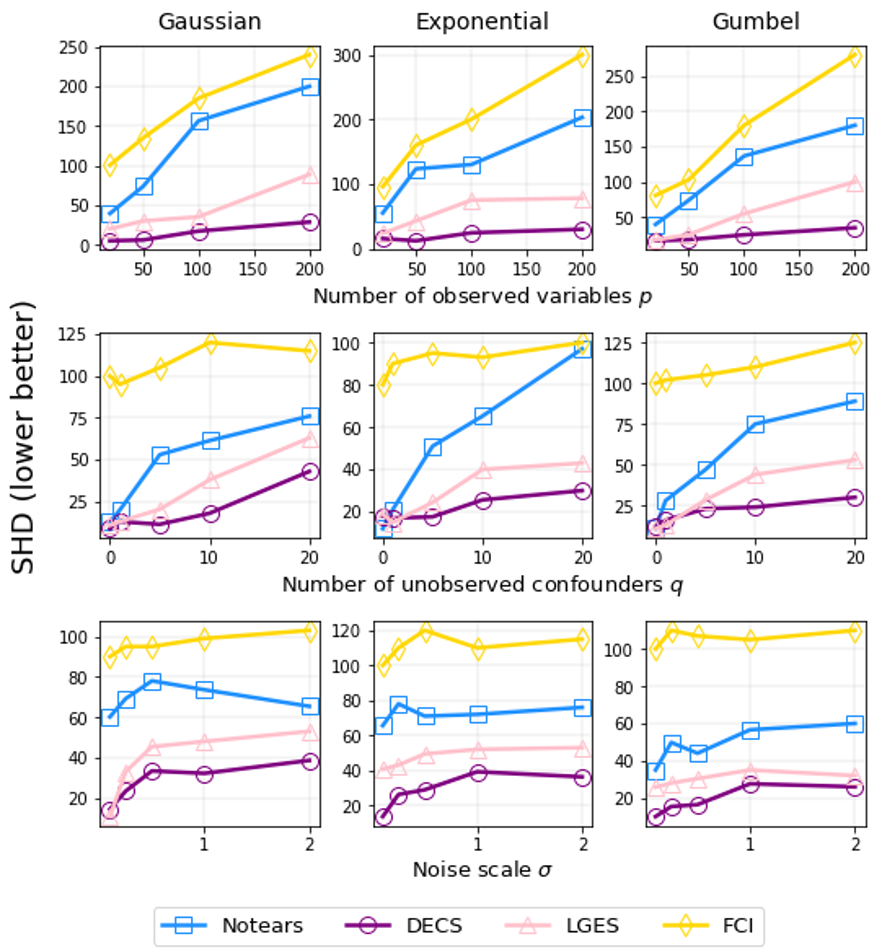} }}%
    \hfill
    \subfloat{{\includegraphics[width=7.4cm]{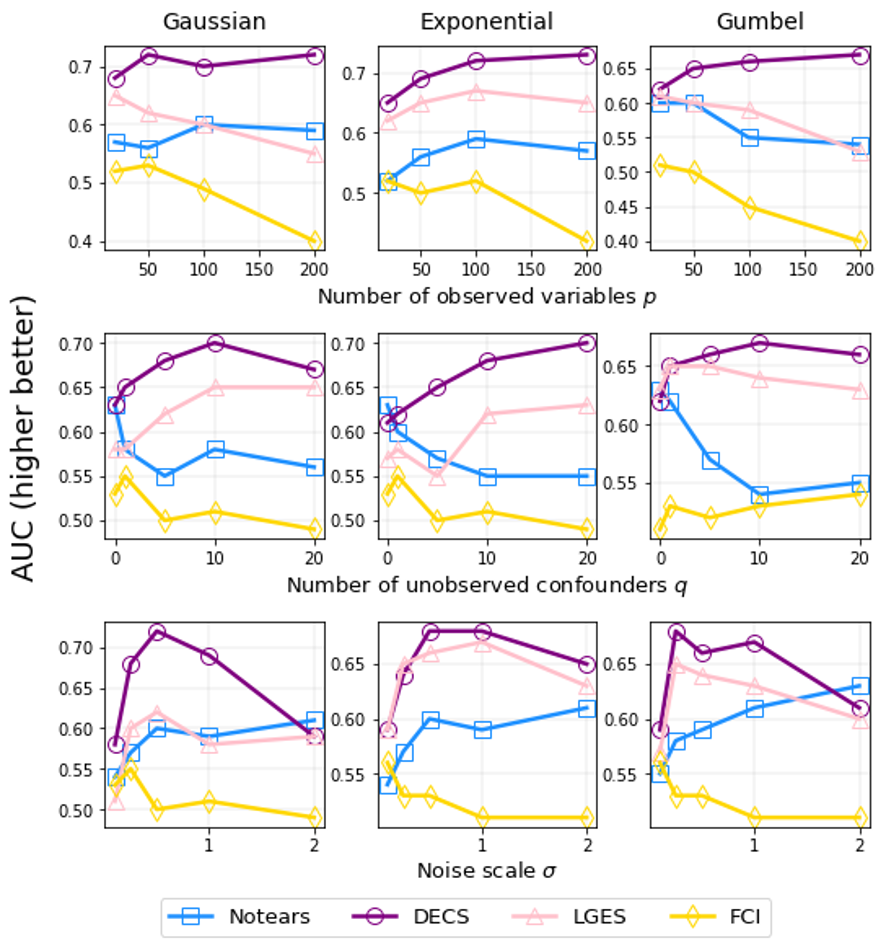} }}%
    \caption{Performance on synthetic experiments. DECS is the proposed approach.}
    \label{fig:perf}
\end{figure*}

\subsection{Results} 

\textbf{(1) The data distribution family.} Each column of Figure \ref{fig:perf} refers to a different data distribution family. We can see that when the Gaussian assumption is satisfied DECS can significantly improve in performance with respect to other methods, especially for relatively high-dimensional graphs (top row). It is interesting however that relative performance does not vary with a change in distribution (Exponential of Gumbel) which suggests that DECS is robust to the underlying noise model. 

\textbf{(2) Dimensionality of observed variables}. In the top row of Figure \ref{fig:perf} we show performance as a function of the dimensionality of the observables. Theoretically, DECS requires  high-dimensional data and we see that outperformance is strongest in this regime (the number of samples here is $100$) although DECS remains competitive otherwise.

\textbf{(3 and 4) Dimensionality and strength of unobserved confounders.} On the middle and bottom rows we vary the dimensionality of unobserved confounders $q$ and strength of confounding (through $\sigma$) respectively. When $q=0$ the system is fully observed. An interesting observation is that Notears and DECS perform similarly which suggests that there is nothing lost by adjusting even without unobserved confounders. As we increase $q$ and the strength of confounding DECS outperforms.

\textbf{(5) Sparse unobserved confounders.} In the Appendix we conduct an experiment to test the sensitivity of DECS with respect to the level of denseness on $B$. The advantage of DECS decreases in this case, though performance remains competitive. 




\begin{table*}
\fontsize{8.5}{9}\selectfont
\begin{center}
\begin{tabular}{|p{1cm}|C{2cm}|C{1.8cm}|C{1.8cm}|C{1.8cm}|C{1.8cm}|}
\cline{3-6}
    \multicolumn{2}{c|}{} & \textbf{E. coli} & \textbf{Starch} & \textbf{Scale-Free} & \textbf{Sachs}\\
    \hline
    \multirow{3}{*}{TPR}&Notears & 0.39 $\pm$ 0.01 &  0.24 $\pm$ 0.01& 0.18 $\pm$ 0.01& 0.58 $\pm$ 0.02\\
                        &LGES & \textbf{0.57 $\pm$ 0.05} & \textbf{0.42 $\pm$ 0.03}& 0.15 $\pm$ 0.01& \textbf{0.66 $\pm$ 0.05}\\
                        &DECS (ours) & 0.34 $\pm$ 0.04 & 0.28 $\pm$ 0.05& \textbf{0.21 $\pm$ 0.05} & 0.33 $\pm$ 0.05\\
    \hline\hline
    \multirow{3}{*}{FDR}&Notears & 0.59 $\pm$ 0.02 & 0.83 $\pm$ 0.05& \textbf{0.12 $\pm$ 0.01} & 0.64 $\pm$ 0.05\\
                        &LGES & 0.66 $\pm$ 0.03 & 0.58 $\pm$ 0.03 & 0.82 $\pm$ 0.05& 0.55 $\pm$ 0.05 \\
                        &DECS (ours) & \textbf{0.32 $\pm$ 0.05} & \textbf{0.50 $\pm$ 0.06}& 0.20 $\pm$ 0.03 & \textbf{0.20} $\pm$ 0.04\\
    \hline\hline
    \multirow{3}{*}{SHD}&Notears & 39.0 $\pm$ 2.25& 192 $\pm$ 10.0 & 40.0 $\pm$ 5.25 & 12.5 $\pm$ 1.50\\
                        &LGES & 51.0 $\pm$ 2.50 & 115 $\pm$ 7.00 & 23.0 $\pm$ 2.00 & 13.0 $\pm$ 1.50 \\
                        &DECS (ours) & \textbf{26.0 $\pm$ 2.00} & \textbf{95.0 $\pm$ 3.00}& \textbf{14.0 $\pm$ 1.25} & \textbf{8.00 $\pm$ 1.00}\\ 
     \hline\hline
    \multirow{3}{*}{AUC}&Notears & 0.60 $\pm$ 0.02 & 0.58 $\pm$ 0.03 & 0.65 $\pm$ 0.03 & \textbf{0.67 $\pm$ 0.03}\\
                        &LGES & 0.62 $\pm$ 0.05 & 0.66 $\pm$ 0.06 & 0.59 $\pm$ 0.05 & 0.66 $\pm$ 0.04 \\
                        &DECS (ours) & \textbf{0.65 $\pm$ 0.03} & \textbf{0.67 $\pm$ 0.04}& \textbf{0.70 $\pm$ 0.05} & 0.65 $\pm$ 0.05\\
\hline
\end{tabular}
\end{center}
\caption{Mean performance and standard deviations over 10 random trials on Ecoli $(n = 100, p=41)$, Starch $(n = 100, p=104)$, Scale-Free $(n = 100, p=200)$ and Sachs $(n = 7466, p=10)$ data. Bold indicates best performance. FCI returns a complete graph in almost all cases and we have omitted it from these results (we have also attempted to use more flexible conditional independence tests \cite{bellot2019conditional}).}
\label{tab:experiments}
\end{table*}

\section{Experiments on Genetic Data}
The study of gene regulatory networks is one area in genomics with the potential to uncover the interactions of molecular regulators that govern the gene expression levels of messenger RNA and proteins: the building blocks of all cell function. We are interested in the problem of recovering the underlying gene expression network from individual samples of gene expression.

\begin{minipage}{.65\textwidth}
\textbf{Problem.} To validate performance on this task, we use a number of gene expression simulation programs that have been constructed based on the behaviour of known simple organisms, all publicly available in the \texttt{bnlearn} R package. We consider gene expression data from an E. coli microorganism \cite{schmidt2004reverse} (\textbf{E. coli}), gene expression data describing starch metabolism of Arabidopsis thaliana \cite{opgen2007correlation} (\textbf{Starch}), data from a scale-free network, found to faithfully describe biological organisms \cite{barabasi1999emergence} (\textbf{Scale-Free}), and protein expression level data from human immune system cells \cite{sachs2005causal} (\textbf{Sachs}). All variables are fully observed in all of the above. We consider inducing unobserved confounding by explicitly removing a number of root nodes in the network after sampling data, see Figure \ref{fig:graph} for an example with the Starch network: data from each of the \textcolor{blue}{blue} nodes in the starch network is omitted thereby inducing spurious correlations among their children. Networks, omitted variables, and other details for all datasets can be found in the Appendix. 
\end{minipage}
\hfill
\begin{minipage}{.32\textwidth}
\begin{figure}[H]
\vspace{-0.4cm}
\captionsetup{skip=5pt}
\centering
\includegraphics[width=1\textwidth]{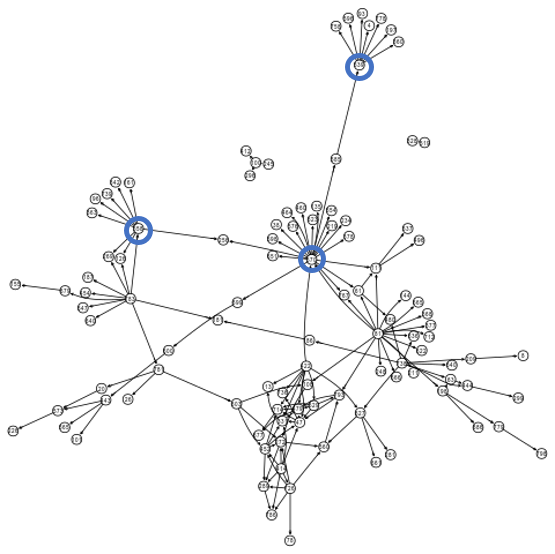} %
    \caption{\textbf{Starch} network.}%
    \label{fig:graph}%
\end{figure}
\end{minipage}

\textbf{Results.} Performance results are given in Table \ref{tab:experiments}. AUC and SHD figures on all datasets show that DECS is competitive on all tasks. We make an additional comparison here considering true positive (TPR) and false discovery (FDR) rates at a threshold chosen for minimum SHD. This comparison is made to show the relatively good false discovery control of DECS even though formal guarantees were not established. On all metrics, and particularly with the AUC that considers performance along the whole threshold spectrum, DECS outperforms in most cases which demonstrates its applicability in realistic genetic data scenarios where knowledge on interactions between genes or gene products are typically not available without interventions.

\subsection{DECS for reproducible discovery}
This section considers reproducibility of causal discovery across environments. If two datasets differ in the distribution of unmeasured variation, correlations between observables vary, and we cannot expect estimates of conventional causal discovery algorithms to be reproducible. This is an important challenge because any two experiments most likely do differ due to changes in environment, data collection practices, among other unmeasured factors. The adjusted adjacency matrix from DECS, by definition removes sources of unmeasured variation from the otherwise biased estimate. We can expect the estimated adjacency matrix to be invariant in theory to changes in distribution of unobserved confounders, and therefore more reproducible and stable across different experiments.

\textbf{Experiment design.} To test this feature, we adopt the scale-free network and construct several datasets while varying the extent of unobserved confounding to simulate different environments\footnote{This experiment considers adjacency matrix recovery but we make additional comparisons on the basis of skeleton recovery with LGES in the Appendix.}. Specifically, we let $X = W X + B H +  E$, where matrices $W$ and $B$, and the distribution $E \sim \mathcal N_p(0,I)$ are fixed, while $H$ is drawn from distributions $\mathcal N(0,\sigma)$ with varying $\sigma$ (one for each environment, drawn at random in the interval $[0.25,2]$).

\begin{minipage}{.65\textwidth}
\textbf{Results.} The problem is to test for agreement between recovered adjacency matrices $W$ in different environments. We report the number of edges that reproduce across different environments in Figure \ref{fig:reproducibility}. Each point on the plot gives the proportion of estimated edges that intersect in any $m$ studies, $m=1,\dots,10$. For instance, approximately $15\%$ of estimated edges (across all 10 environments) intersect in all 10 environments for DECS whereas only $1\%$ do for Notears. This shows that adjusting for unobserved confounding improves the reproducibility of causal discovery.
\end{minipage}
\hfill
\begin{minipage}{.33\textwidth}
\begin{figure}[H]
\vspace{-0.4cm}
\captionsetup{skip=5pt}
\centering
\includegraphics[width=0.9\textwidth]{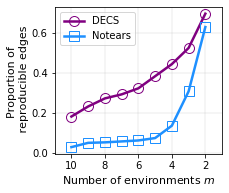}
    \caption{Reproducibility.}%
    \label{fig:reproducibility}
\end{figure}
\end{minipage}

\section{Conclusions}
This paper develops a score-based causal discovery algorithm in the presence of dense unobserved confounding (unobserved variables with a widespread effect on observed ones). The argument considers properties of the spectrum of the data matrix that allows DAG learning (directed edges among observed variables) in the presence of dense confounding to be expressed as a continuous optimization problem. Solutions to this problem have guarantees on the true positive rate in the high-dimensional regime, the resulting score-based problem is much simpler to implement than independence-based alternatives and it outperforms empirically across a range of different experiments.

One may extend the proposed approach to model more general structural models. Specifically, structural models not constrained by a specific data distribution family or functional relationships between variables. One may consider as an extension optimization problems of the form,
\begin{align}
    \underset{f\in\mathcal F}{\text{arg min}}\hspace{0.3cm} \frac{1}{n}\sum_{i=1}^n\mathcal L \left(\mathbf{\tilde x}_i, f(\mathbf{\tilde x}_i)\right) + \rho_{\lambda}(f)
\end{align}
where $\mathcal F$ is a more general space of functions $f:\mathbb R^p \rightarrow \mathbb R^p$ that defines the causal structure in the data through its partial derivatives with respect to its arguments, $\mathcal L:\mathbb R^{p \times p} \rightarrow \mathbb R$ is a loss function (that may be chosen to model other data types, such as binary or count data) and $\rho_{\lambda}(f)$ is a regularization term that includes the acyclicity constraint. In this case, it takes a different form but may be computed for large classes of functions by considering norms on partial derivatives as in \cite{zheng2020learning} and has already been shown to be successful for non-linear models in the fully observed setting.

There is scope as well for considering other adjustment frameworks that control for the influence of unobserved confounding. For instance, using different problem-dependent eigenvalue thresholds in the adjusted data matrix or by optimizing simultaneously for matrices $W$ and $B$ in the linear structural model with an $l_1$ and $l_2$ penalty respectively as \cite{chernozhukov2017lava} considered in the regression setting.

\section*{Acknowledgements}
This work was supported by the Alan Turing Institute under the EPSRC grant EP/N510129/1, the ONR and the NSF grants number 1462245 and number 1533983.

\bibliography{bibliography}
\bibliographystyle{plain}

\newpage

\appendix
\onecolumn
{\Large \textbf{Appendix}}
\\\\
This Appendix provides additional details to supplement the main body of this paper. It is outlined as follows:
\begin{itemize}[leftmargin=*]
    \item Section \ref{sec_proof_DECS} contains proofs.
    \item Section \ref{sec_synthetic_experiments_DECS} includes further simulations and details of the synthetic experiments and implementations.
    \begin{itemize}
        \item Section \ref{sec_details_DECS} gives details of the synthetic experiments.
        \item Section \ref{sec_sparsity_DECS} includes an experiment analysing performance with sparse unobserved confounding.
        \item Section \ref{sec_error_DECS} analyses the recovery of the exact weighted adjacency matrix with synthetic simulations.
        \item Section \ref{sec_reproducibility_DECS} gives further reproducibility experiments on skeleton recovery.
    \end{itemize}
    \item Section \ref{sec_genetic_experiment_DECS} gives details of the (semi-synthetic) genetic experiments.
\end{itemize}



\section{Proofs}
\label{sec_proof_DECS}

\subsection{Proof of Lemma 1}

\textbf{Lemma 1.} \textit{Assume that $H$ is univariate. Then, the principal components of $X$ are approximately aligned with the columns of $B$ and approximately aligned with each row of the confounded contribution $C$.}

\textit{Proof.} Assume that $H$ is a standard univariate Gaussian random variable. Let $\Sigma_E:=\Cov(E)\approx I$. $B\in\mathbb R^p$ (a column vector since $H$ is univariate) tends to be approximately aligned with $\Cov(X) = (I-W)^{-1} (BB^T+\Sigma_E) (I-W)^{-T}$ since $B$  is the largest eigenvector of $(BB^T+I)$ and $W$ is sparse. The $i$-th column of the perturbation $C$ is defined as,
\begin{align}
    c_i = \Cov(X)^{-1}\Cov(X,H)B = B_i(I-W)^T(BB^T+\Sigma_E)^{-1}B.
\end{align}
By the Sherman-Morrison formula,
\begin{align}
    (BB^T+\Sigma_E)^{-1} = \Sigma_E^{-1} - \frac{\Sigma_E^{-1}BB^T\Sigma_E^{-1}}{1+B^T\Sigma_E^{-1}B}.
\end{align}
If we assume $\Sigma_E\approx I$ then it follows that $(BB^T+\Sigma_E)^{-1}B \approx \lambda B$ where $\lambda$ is a scalar because $B$ is an eigenvector of $I - \frac{BB^T}{1+B^TB}$. With $W$ sparse thus we may then extend this approximation to say that $C_i$ is approximately aligned with large eigenvectors of $\Cov(X)$ for each $i$. This can be made precise with perturbation theory of Hermitian matrices. 

To quantify the approximation of the direction of eigenvectors of $(I-W)^{-1} (BB^T+I) (I-W)^{-T}$ by those of $BB^T+I$ we can apply Theorem 4.1 in \cite{li1998relative} writing $U$ and $\tilde U$ for the matrices of eigenvectors of $BB^T+I$ and $(I-W)^{-1} (BB^T+I) (I-W)^{-T}$ respectively. We have that,
\begin{align}
    \|\sin\Theta(U,\tilde U)\|_F \leq \frac{1}{\sigma}\sqrt{\|WU\|^2_F + \|I-(I-W)^{-1}U\|^2_F},
\end{align}
where $\Theta(\cdot,\cdot)$ is the canonical angle between column spaces defined e.g. in equation 2.5 in \cite{li1998relative}, where $\sigma$ is a scalar related to the minimum distance between eigenvalues of $BB^T+I$.

\subsection{Proof of Theorem 1}

We begin by recalling the adjusted regression model that we seek to analyse.
\begin{align}
\label{adjusted_model}
    &FX =  FX (W+C) + F \bar E \qquad
    \Rightarrow\qquad  \tilde X =  \tilde X (W+C) + \tilde E
\end{align}
Let us write $\tilde \Sigma = \Cov(\tilde X)$ for the covariance matrix of $\tilde X$. Even for a good choice of $F$ that balances between a well behaved error term $\tilde E = F \bar E$, well behaved design matrix $\tilde X$ and well behaved perturbation term $\tilde X C$ tending to zero, $W$ is not necessarily uniquely identifiable. The map between the observed covariance $\tilde \Sigma$ and the pair of causal adjacency matrix $W$ and error covariance $\tilde \Sigma_E = \Cov(\tilde E)$ is not necessarily unique. To avoid issues of identifiability, recent work \cite{aragam2015learning} defines minimum-trace DAGs $W_{\min}$,
\begin{align}
    (W_{\min}, \Sigma_{\min}) \in \text{arg min} \{Tr(\tilde\Sigma): (W, \tilde\Sigma_E) \in \mathcal D \}
\end{align}
where $\mathcal D$ denotes all pairs $(W, \tilde\Sigma_E)$ that exhibit a data covariance indistinguishable from that observed. Minimum-trace DAGs themselves are not necessarily unique in general but for the purposes of the results presented here we will assume it to be unique for good choices of $F$ that shrink the spurious signal without altering the causal signal too much. We note that extensions exist for unidentifiable case \cite{aragam2019globally}, in which case penalized score optimization can be shown to converge to a sparse representative within the class of minimum-trace DAGs but leave this investigation in the presence of unobserved confounding to future work.

Our objective is to control the likelihood of the following failure event,
\begin{align}
    \{\text{supp}(W_{\min}) \subsetneq \text{supp}(\widehat W)\},
\end{align}
where $\widehat W$ is the solution to the constrained, penalized optimization program,
\begin{align}
\label{app_optimization}
    \widehat W \in \underset{W \in \mathbb D}{\text{argmin}}\hspace{0.3cm} \mathcal S(W; \mathbf X), \qquad
    \mathcal S(W; \mathbf X):=\frac{1}{2n} || \tilde{\mathbf X} - \tilde{\mathbf X} W ||^2_F + \lambda ||W||_1.
\end{align}
This can be done by reducing the analysis of $\widehat W$ to a family of neighbourhood regression problems \cite{aragam2015learning,aragam2019globally}. There are two key steps: 

\begin{enumerate}[leftmargin=*]
    \item First showing that $\widehat W$ is equivalent to solving a series of $p$ regression problems given by,
    \begin{align}
    \label{app_cevic}
        \underset{\mathbf w_i \in\mathbb R^p,\hspace{0.1cm}  \text{supp}(\mathbf w_i)\subset S}{\text{arg min}}\hspace{0.3cm} \frac{1}{2n} || \tilde{\mathbf{X}}_i -  \tilde{\mathbf X} \mathbf w_i ||^2_2 + \lambda ||\mathbf w_    i||_1
    \end{align}
    as defined in the main body of this paper.
    \item And second, controlling for the error in estimation in each of these neighbourhood problems for all subsets of covariates, or neighbourhoods given by $S$. 
\end{enumerate} 
    
\subsubsection{First step}
The first step is a consequence of how the least squares loss and regularizer factor. This allows to formally establish the equivalence between the DAG problem and neighbourhood regression, and is justified by Lemma B.1. in \cite{aragam2019globally}. This is similar to undirected models, for which the analysis can be reduced to $p$ different regression problems, namely the regression of $X_j$ onto $X_{-j}$. Unfortunately, for DAGs, there are $p2^p$ possible regression problems (the regression of $X_j$ onto any subset of other variables $S$), which quickly become intractable to control uniformly. In the identifiable case, we can constrain ourselves to control over sets $S$ that are consistent with a superstructure $G$ of the underlying graph, i.e. we must only control over those adjacency matrices that are sub-graphs of $G$ (e.g. the moral graph of a DAG is an example of superstructure). \cite{aragam2019globally} then show a uniform concentration bound for the score function restricted to a consistent superstructure and use this result to show that any estimated $\hat W$ has the same topological sort as $W_{\min}$. This topological sort identifies candidate parent sets for each node $X_j$, and reduces the problem to control over $p$ regression problems, which is substantially lower than $p2^p$ problems. 

These steps rely on the model distribution, independence of the error term in (\ref{adjusted_model}), and the properties of minimum-trace DAGs, and are given as a sequence of Lemmas and Propositions in Appendix B in \cite{aragam2019globally}. All proofs (and prior conditions for the applicability of each statement) therein hold for our model without modification since the distribution family is preserved under deterministic transformations of both sides of the model equation, and the independence of error terms holds by construction of the matrix $C$ and Gaussianity. We refer the reader to these references for a detailed derivation of each of these steps.

\subsubsection{Second step}
The second point differs from \cite{aragam2019globally}. It holds that the optimization program (\ref{app_optimization}) can be reduced to a collection of local regression problems, but in our case each regression problem is defined as (\ref{app_cevic}) rather than the conventional un-adjusted lasso. For this problem, as mentioned, a good choice of $F$ needs to find a balance between a well behaved error term $\tilde E = F \bar E$, well behaved design matrix $\tilde X$ and well behaved perturbation term $\tilde X C$. These conditions can be articulated in three assumptions on the adjusted program. 
\begin{itemize}[leftmargin=*]
    \item Assumption 1 from the main body of this work: we assume $\sigma_{\min}(B) = \mathcal O(\sqrt{p})$ which implies $\sigma_{\min}(\Cov(X,H)) = \sigma_{\min}((I-W)^{-1}B) = \mathcal O(\sqrt{p})$: the largest singular value of the $(p \times q)$ covariance matrix of $(X, H)$ is of the order $\sqrt{p}$, which is a consequence of denseness of unobserved confounding since the smallest singular value of Gaussian, i.e. dense, random matrices is lower bounded by a term of the order of $\sqrt{p}$ \cite{rudelson2009smallest}.
    \item We assume that $\tilde d_{n/2} = \mathcal O(\sqrt{p})$: the median value of the singular values of $\Cov(X)$ (and maximum singular value of $\tilde X$) is of the order $\sqrt{p}$, with high probability.
    \item We assume that the compatibility constant $\phi_M$ of $M := n^{-1}\mathbf{\tilde{X}}^T \mathbf{\tilde{X}}$ is of the same order as the minimal singular value of $X$: there exists a constant $c > 0$ such that $P(\phi_M^2 / \sigma_{\min}(\Cov(X)) > c) \rightarrow 0$. The compatibility constant is a kind of restricted eigenvalue condition and is common in the model selection literature, see e.g. \cite{cevid2018spectral}. For a square matrix $M$ it is defined as,
    \begin{align}
        \phi_M := \underset{||\alpha||_1 \leq 5||\alpha_S||_1}{\text{inf}}\frac{\sqrt{\alpha^T M \alpha}}{||\alpha_S||_1/\sqrt{s}},
    \end{align} 
    where $S$ is the support set of $\mathbf w_i$, $s$ is the size of $S$ and $\alpha_S$ is a vector consisting only of the components of $\alpha$ which are in $S$.
\end{itemize}

These conditions, by Theorem 1 in \cite{cevid2018spectral}, are sufficient for the error in estimation of $\mathbf w_i$ with the program (\ref{app_cevic}) to be bounded in $l_1$ norm by a factor of order,
\begin{align}
    \mathcal O\left(\frac{\sigma_i s}{\sigma_{\min}(\tilde\Sigma_E)}\sqrt{\log p / n}\right).
\end{align}
$\sigma_i$ is the standard deviation of $\tilde E_i$ and $s$ is the size of the support of $\mathbf w_i$. 

We now assumed an additional beta-min condition, i.e. a condition minimum strength on the signal of causal coefficients,
\begin{align*}
    \text{min}(|w|: w \in \text{supp}(W_{\text{min}}))\gtrsim \sigma \sqrt{\log p / n},
\end{align*} 
where we have written $a \gtrsim b$ to mean that $a \geq C \cdot b$ for some constant $C > 0$, and $\sigma = \frac{\max_i(\sigma_i) s}{\sigma_{min}(\tilde\Sigma_E)}$. 

Control over events of the form $\{\text{supp}(\mathbf{w}_{i}) \subsetneq \text{supp}(\mathbf{\hat w}_i)\}$, which is the second key step, then follows from the following inequality,
\begin{align}
    || \mathbf{\hat w}_i - \mathbf w_i ||_1 \leq ||  \mathbf{\hat w}_i - \mathbf w_i ||_{\infty}
\end{align}
It follows that $\text{supp}(\mathbf w_i) \subseteq\text{supp}(\mathbf{\hat w}_i)$ as long as $\text{min}(|w|: \text{supp}(w) \in \text{supp}(\mathbf w_i))\gtrsim \sigma \sqrt{\log p / n}$ with high probability. If not, we could find a $j \in \text{supp}(\mathbf w_i)$ with $j \notin \text{supp}(\mathbf{\hat w}_i)$ such that $|\hat w_{ij} - w_{ij}| = |\hat w_{ij}| \gtrsim \sigma \sqrt{\log p / n}$, which leads to a contradiction. Here $w_{ij}$ is the $j$-th element of the vector $\mathbf{w}_i$.

Finally, control over false positives $\{\text{supp}(\mathbf w_i) \nsubseteq\text{supp}(\mathbf{\hat w}_i)\}$ in each neighbourhood regression problem implies control over events $\{\text{supp}(W_{\min}) \nsubseteq \text{supp}(\widehat W)\}$ in DAG estimation by a uniform bound over the control ensured in the $p$ distinct neighbourhood regression problems, and is technically justified by point (b) in Lemma B.1 in \cite{aragam2019globally}, that ensures that $\widehat W$ is the unique solution to (\ref{app_optimization}) if and only if $\mathbf{\hat w_i} = [\widehat W]_{\cdot i}$ is the unique solution to (\ref{app_cevic}).



\section{Details on synthetic experiments}
\label{sec_synthetic_experiments_DECS}

\subsection{Simulations, metrics and implementation}
\label{sec_details_DECS}

In the main body of this paper, we consider one main synthetic network model:
\begin{itemize}[leftmargin=*]
    \item Erd\"os–R\'enyi graph models. These are generated by adding edges independently with equal probability $r = \frac{2e}{p^2-p}$, where $e$ is the expected number of edges in the resulting graph. For each $p$-node graph, we simulate graphs with $e$ equal to $p$.
\end{itemize}

Based on the DAG sampled from this graph model, we assign edge weights sampled independently from Uniform$([-2, -0.5] \cup [0.5, 2])$ to construct the weighted adjacency matrix $W \in\mathbb R^{d\times d}$. The observational data is then generated according to the linear confounded DAG model with different graph sizes, and additive noise types:
\begin{itemize}[leftmargin=*]
    \item Gaussian. $H_i,E_j \sim \mathcal N (0, 1), \quad i = 1, \dots, q,\quad j = 1, \dots, p.$
    \item Exponential. $H_i,E_j \sim \text{Exp} (1),\quad i = 1, \dots, q, \quad j = 1, \dots, p.$
    \item Gumbel. $H_i,E_j \sim \text{Gumbel} (0, 1),\quad i = 1, \dots, q,\quad j = 1, \dots, p.$

\end{itemize}
In each synthetic experiment we generate $n = 100$ samples for each of these settings. For experiments considering performance as a function of varying dimensionality $p$ of $X$, we fixed $q=10$ and $\sigma = 0.2$. For experiments considering varying dimensionality $q$ of $H$, we fixed $p=20$ and $\sigma = 0.2$. For experiments considering varying $\sigma$, we fixed $p=20$ and $q=10$.

We evaluate the estimated graphs using four different metrics:
\begin{itemize}[leftmargin=*]
    \item Structural Hamming Distance (SHD) indicates the number of edge additions, deletions, and reversals in order to transform the estimated graph into the ground truth DAG.
    \item True Positive Rate (TPR) measures the proportion of actual positive edges that are correctly identified as such.
    \item False Discovery Rate (FDR) measures the proportion of false discoveries among the estimated edges.
    \item The Area Under the ROC Curve (AUC) measures the area under a plot of the TPR as a function of FDR as te threshold for determining presence / absence of edges is varied.
    \item The $l_2$ loss in the recovery of adjacency matrices $||\widehat W - W||_2^2 / p$.
\end{itemize}

We use the following implementations for baseline algorithms.
\begin{itemize}[leftmargin=*]
    \item FCI was implemented through the \texttt{pcalg} R package with a Gaussian conditional independence test.
    \item LGES was implemented with hyperparameters chosen by cross validation following the author's implementation at \texttt{https://github.com/benjaminfrot/lrpsadmm/}.
    \item NOTEARS. We use the variant with $l_1$ regularization chosen by cross-validation. The code is available at the author’s GitHub repository
    \texttt{https://github.com/xunzheng/notears}.
\end{itemize}

\begin{figure}[t]%
\centering
    \includegraphics[width=8cm]{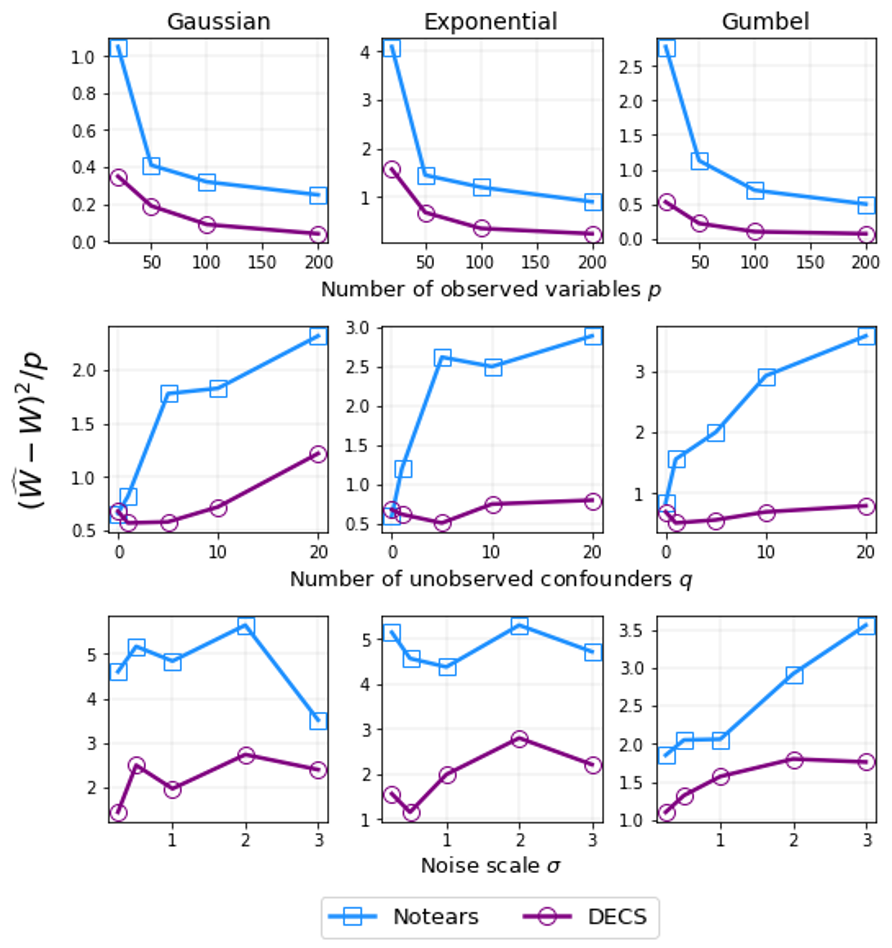} %
    \caption{Performance on the recovery of the weighted adjacency matrix.}%
    \label{fig:error}%
\end{figure}

\subsubsection{How to compute AUC and SHD on the different baselines} 
We have mentioned that all comparisons are made using estimated skeletons. The AUC considers a range of precision / recall values estimated with different parameters to determine the presence / absence of edges. 

\begin{itemize}[leftmargin=*]
    \item For DECS and Notears this computation is straightforward as both return weighted adjacency matrices and one obtains a skeleton by choosing different thresholds on the estimated weights to determine presence / absence of edges.
    \item For LGES the strategy is different as it does not return weighted adjacency matrices. The equivalence class of LGES is computed using the BIC and we obtain a range of precision / recall values by considering a range of penalties on the strength of the BIC regularization, as done by the authors in \cite{frot2017robust}. 
    \item FCI uses independence tests to recover the skeleton and thus requires a threshold for significance, precision / recall values are obtained by varying this threshold.
\end{itemize}

\subsection{Further experiments with sparse unobserved confounding}
\label{sec_sparsity_DECS}
We conduct in this section an empirical investigation on the sensitivity of DECS with respect to the level of denseness on $B$. Sparse unobserved confounding render the spurious contributions to the adjacency matrix indistinguishable from the true causal signal. We consider $B$ to be drawn as $W$ in the data generating mechanism, i.e. a DAG with a specified number of edges $e$ (fewer edges implying sparser unobserved confounding contribution). 

We evaluate all algorithms on the Gaussian model with Erd\"os–R\'enyi and $p=20$ nodes with the difference that $B$ is drawn as $W$ with $e$ edges (recall that $W$ has fixed $e=20$ edges). 

As can be seen in Table \ref{tab:supp_experiments}, with decreasing number of non-zero entries in $B$, that is increasing sparsity, the advantage of DECS decreases, though performance remains competitive.

\begin{table}[H]
\fontsize{9.5}{9.5}\selectfont
\begin{center}
\begin{tabular}{p{1.6cm}C{1.4cm}C{1.4cm}C{1.4cm}C{1.4cm}}
\toprule
     & \textbf{20} & \textbf{50} & \textbf{100} & \textbf{200}\\
     \midrule
    DECS & $62 \pm 6.0$ & $53 \pm 7.3$ & $50 \pm 6.2$ & $46 \pm 5.9$\\
    NOTEARS & $68 \pm 3.9$ & $71 \pm 2.5$ & $70 \pm 2.8$ & $75 \pm 2.0$\\
    LGES & $56 \pm 3.3$ & $60 \pm 5.1$ & $58 \pm 6.0$ & $53 \pm 3.4$\\
    FCI & $67 \pm 6.2$ & $85 \pm 3.7$ & $85 \pm 5.0$ & $95 \pm 5.7$\\
\bottomrule
\end{tabular}
\end{center}
\caption{SHD as a function of the number of non-zero entries in $B$}
\label{tab:supp_experiments}
\end{table}

\subsection{Further experiments using adjacency matrix error}
\label{sec_error_DECS}
In the main body of this paper we tested performance on undirected graphs to allow for comparisons across algorithms with different outputs. Here we consider recovery performance of the original weighted adjacency matrix $W$ used to generate the data. Comparisons are made with Notears which is the only method that returns a weighted adjacency matrix although it does not account for unobserved confounding. This experiment thus served to show that adjusting for unobserved confounding can significantly improve upon the same algorithm without adjustments.

We follow the same experimental set-up as in the main body of this paper and report results in Figure \ref{fig:error}.

\subsection{Further reproducibility experiments on skeleton recovery}
\label{sec_reproducibility_DECS}
In the main body of this paper we tested for the reproducibility of causal discovery in different environments shifted by the distribution of unobserved confounders. In this section we consider the exact same set-up but test instead for skeleton recovery to be able to make comparisons with LGES.

\begin{figure}[H]%
\centering
    \includegraphics[width=5cm]{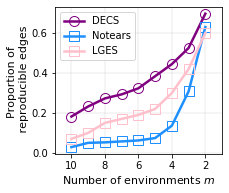} %
    \caption{Reproducibility experiments on skeleton recovery. Higher values for larger number of environments indicate higher levels of reproducibility. DECS is the proposed approach.}%
    \label{fig:app_reproducibility}%
\end{figure}

Results are given in Figure \ref{fig:app_reproducibility}. The results show that DECS returns a skeleton which is more reproducible across environments. For instance, approximately $20\%$ of estimated edges in the skeleton (across all 10 environments) intersect in all 10 environments for DECS whereas only $7\%$ and $3\%$ do for LGES and Notears respectively.

\newpage
\section{Details on Genetic (semi-synthetic) data}
\label{sec_genetic_experiment_DECS}

\begin{itemize}[leftmargin=*]
    \item The \textbf{Scale Free (SF)} graph is simulated using the Barab\'asi-Albert model \cite{barabasi1999emergence}, which is based on the preferential attachment process, with nodes being added sequentially. In particular, $1$ edge is added each time between the new node and existing nodes. Scale-free graphs are popular since they exhibit topological properties similar to real-world networks such as gene networks, social networks, and the internet. Once the network $G$ is sampled we draw edge weights and data following the Erd\"os-R\'enyi data generating process with $n=100, p=200, q=10, \sigma = 0.2$.
    \item The \textbf{E. coli} network describes the expression of protein coding genes of the E. coli microorganism under stress, in an experiment conducted by \cite{schmidt2004reverse}. The available data of $100$ samples of $46$ genes was sampled from a Gaussian model, as described in the \texttt{bnlearn} R package.
    \item The \textbf{Starch} network simulates gene expression expression interaction resulting from an experiment investigating the impact of the diurnal cycle on the starch metabolism of Arabidopsis thaliana \cite{opgen2007correlation}. This gene network and data contains $107$ genes, $150$ edges and $100$ samples and represents an example of a high-dimensional causal discovery problem. It is available in the \texttt{bnlearn} R package.
    \item The \textbf{Sachs} dataset consists of $n = 7466$ measurements of expression levels of proteins and phospholipids in human immune system cells for $p = 11$ cell types \cite{sachs2005causal}. It is widely used as a benchmark for causal discovery as it comes with a consensus network that is accepted by the biological community. It is available in the \texttt{bnlearn} R package.
\end{itemize}

\begin{figure*}%
    \centering
    \subfloat[Starch network]{\includegraphics[width=4cm]{Figures/arth.png} }%
    \quad
    \subfloat[E. coli network]{\includegraphics[width=4cm]{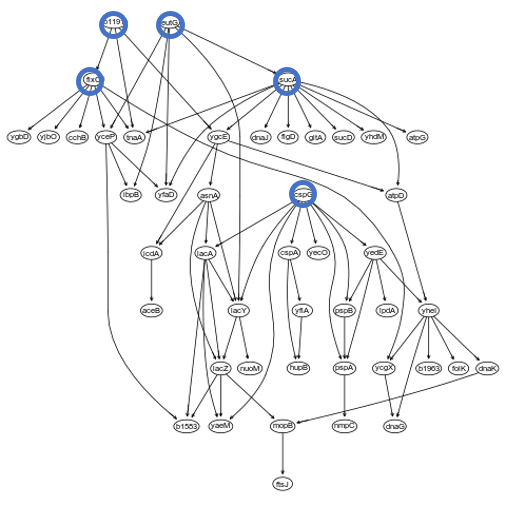} }
    \quad
    \subfloat[Sachs network]{\includegraphics[width=4cm]{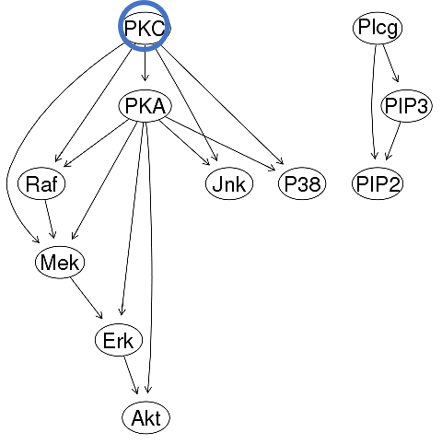} }
    \caption{Networks and omitted variables considered in the genetic data experiments.}
    \label{fig:networks}
\end{figure*}

We give illustrations of the real networks, together with omitted nodes in Figure \ref{fig:networks}. Variables in blue are root nodes omitted from the available data to induce unobserved confounding among children, and thus simulate a scenario of incomplete system of variables as would be expected in real applications.

\end{document}